\documentclass[10pt,twocolumn,letterpaper]{article}

\usepackage[pagenumbers]{wacv} 

\usepackage{algorithm}
\usepackage{algpseudocode}
\usepackage{subcaption}
\usepackage{graphicx}
 \usepackage{multirow}

\definecolor{wacvblue}{rgb}{0.21,0.49,0.74}
\usepackage[pagebackref,breaklinks,colorlinks,allcolors=wacvblue]{hyperref}

\def\wacvPaperID{1102} 
\def\confName{WACV}
\def\confYear{2026}

\title{3-Dimensional CryoEM
Pose Estimation and Shift Correction
Pipeline}

\author{Kaishva Chintan Shah\\
Indian Institute of Technology, Bombay\\
Mumbai, India\\
\and
Virajith Boddapati\\Indian Institute of Technology, Bombay\\
Mumbai, India\\
\and Karthik S. Gurumoorthy\\
Walmart Global Tech\\
Bengaluru, India\\
\and
Sandip Kaledhonkar\\Indian Institute of Technology, Bombay\\
Mumbai, India\\ 
\and Ajit Rajwade\\
Indian Institute of Technology, Bombay\\
Mumbai, India\\}

\begin{document}
\maketitle
\begin{abstract}
The aim of cryo-electron microscopy (cryo-EM) is to estimate the 3D density map of a macromolecule from thousands of noisy 2D projection images (called particles), each acquired in a different orientation and with a possible 2D shift. Accurate pose estimation and shift correction are key challenges in cryo-EM due to the very low SNR, which directly impacts the fidelity of 3D reconstructions. We present an approach for pose estimation in cryo-EM that leverages multi-dimensional scaling (MDS) techniques in a robust manner to estimate the 3D rotation matrix of each particle from pairs of dihedral angles. We express the rotation matrix in the form of an axis of rotation and a unit vector in the plane perpendicular to the axis. The technique leverages the concept of common lines in 3D reconstruction from projections. However, common line estimation is ridden with large errors due to the very low SNR of cryo-EM projection images. To address this challenge, we introduce two complementary components: (i) a robust joint optimization framework for pose estimation based on an $\ell_1$-norm objective or a similar robust norm, which simultaneously estimates rotation axes and in-plane vectors while exactly enforcing unit norm and orthogonality constraints via projected coordinate descent; and (ii) an iterative shift correction algorithm that estimates consistent in-plane translations through a global least-squares formulation. While prior approaches have leveraged such embeddings and common-line geometry for orientation recovery, existing formulations typically rely on $\ell_2$-based objectives that are sensitive to noise, and enforce geometric constraints only approximately. These choices, combined with a sequential pipeline structure, can lead to compounding errors and suboptimal reconstructions in low-SNR regimes. Our pipeline consistently outperforms prior methods in both Euler angle accuracy and reconstruction fidelity, as measured by the Fourier Shell Correlation (FSC). 
\end{abstract}
\section{Introduction}
Cryo-electron microscopy (cryo-EM) has revolutionized structural biology by enabling near-atomic resolution imaging of biological macromolecules \cite{Frank2018}. It circumvents the need for crystallization, making it especially valuable for large or flexible complexes. In cryo-EM, thousands of copies of a macromolecule (e.g., virus, ribosome, etc.) are put into a test-tube containing a solvent. The contents of the test-tube are poured onto a slide and frozen. The slide is imaged by a cryo-electron microscope which shoots electron beams vertically downwards on the slide and acquires an image called a `micrograph'. The micrograph contains the 2D projection images (called `particles') of each macromolecule copy in a random and unknown orientation, against a noisy background.
The particles are extracted from the noisy background via a technique called particle picking, which typically uses modern machine learning approaches \cite{Xu2025}. The core computational challenge lies in reconstructing the 3D structure from thousands of noisy particles, each of which has an unknown 3D orientation (a rotation matrix) and unknown 2D shift\footnote{Any shift in the $Z$ direction has no effect on the 2D projection.}. The unknown 2D shift is due to possible errors in locating the center of the particle during particle picking. Accurate pose estimation, i.e., recovering both orientation and in-plane rotation of each projection, is essential, as errors at this stage directly impact the fidelity of the final reconstruction.
Despite advances in computational methods, pose estimation remains challenging due to the extremely low signal-to-noise ratio (SNR) in projections and unknown translational shifts. 

If the underlying 3D structure is $f(x,y,z)$, then the $i$th projection image ($1\leq i\leq n$), corresponding to unknown 3D rotation matrix $\boldsymbol{R}_i \in \mathcal{SO}(3)$ and unknown shift $\boldsymbol{s}_i := (\Delta x_i,\Delta y_i,0)$, is given by:
\begin{equation}
h_i = \int f(\boldsymbol{R}^{\top}_i \boldsymbol{p} - \boldsymbol{s}_i) dz,    
\label{eq:projection}
\end{equation}
where $\boldsymbol{p} := (x,y,z)^{\top}$ is a spatial coordinate. The planes corresponding to projection images in different poses intersect in a single line called the `common line'. These common lines between different pairs of projection images carry information which is useful in estimating the unknown orientations $\{\boldsymbol{R}_i\}_{i=1}^n$ as has been exploited in several papers such as \cite{Vanheel_1981,singer2010detecting,muller2024algebraic}. Given a pair of projection images $h_i, h_j$ from different poses where $i \neq j$, the common line is essentially computed by finding the most similar pair of lines passing through the center of $h_i$ and $h_j$. Given a triple of projection images $h_i, h_j, h_k$ in different non-coplanar poses, the common lines between each pair helps determine the orientation of each projection image relative to the other via a method called angular reconstitution \cite{van1987angular}. However the low SNR of the particle images induces huge errors in the process of common line detection which adversely affects orientation estimation. Recent geometric approaches use global optimization over angular constraints via frameworks such as spherical embedding \cite{wang2024orientation,lu2022} or synchronization matrices \cite{Shkolnisky2012_sync,wang2013_lud} to estimate projection orientations. However, these methods minimize $\ell_2$ losses that are sensitive to outliers, or enforce orthogonality of the rotation matrix only approximately. Moreover, these methods assume that the projections are pre-aligned (that is, they are centered and hence there are no shifts), limiting robustness in real-world scenarios. As we demonstrate later, our technique yields superior pose estimation results as compared to existing techniques such as \cite{lu2022,wang2013_lud,Shkolnisky2012_sync}.

There exist other methods such as RELION \cite{scheres2012relion} or Cryo-Sparc \cite{punjani2016building} which estimate the unknown structure $f$ and the pose parameters $\{\boldsymbol{R}_i, \boldsymbol{s}_i\}_{i=1}^n$ iteratively in an expectation-maximization (ExM) framework which can be time-consuming. However the emphasis in our work is on techniques that \emph{efficiently} determine pose parameters first and use these as input for image reconstruction without the added feedback loop. This is along the lines of the ASPIRE toolbox for cryo-EM\footnote{\url{http://spr.math.princeton.edu/}}. The advantage of such an approach is its simplicity. We also note that such an approach can be easily incorporated within an iterative ExM framework as a very principled and efficient initialization. 

The main contributions of our paper are:
\begin{itemize}[noitemsep]
\item A joint pose estimation framework using $\ell_1$-norm minimization over both dihedral and in-plane angles under hard (as they should be) orthogonality constraints.
\item A shift refinement method that enforces global consistency across projections via sparse linear equations derived from common lines.
\item Empirical validation on synthetic and real cryo-EM datasets demonstrating improved orientation recovery and reconstruction quality over state-of-the-art baselines.
\end{itemize}
Our approach significantly improves Euler angle estimation accuracy and structural fidelity across simulated datasets, as evidenced by sharper Fourier Shell Correlation (FSC) curves \cite{van2005fourier} and reduced angular errors.

\section{Method}
\subsection{Common-line detection and voting method}
In ideal conditions, the Central Section Theorem states that the 2D Fourier transform of a projection image corresponds to a central slice through the 3D Fourier transform of the underlying structure, orthogonal to the projection direction. Consequently, the Fourier transforms of any two projections intersect along a common line. We estimate these common lines between image pairs via normalized cross-correlation~\cite{Vanheel_1981}.

Following this, dihedral angles between image pairs are estimated using the probabilistic voting strategy introduced in \cite{Singer2010}. For a given pair of projection images $(h_i, h_j)$, this method aggregates dihedral angle estimates $\theta_{ij}^{(k)}$ derived by involving a third image $h_k$ where $k \ne i,j$. Each choice of $k$ contributes a potentially noisy angle estimate. These estimates are combined into a smooth histogram using Gaussian kernels, with the peak of the histogram determining the final dihedral angle:
\begin{equation}
\theta_{ij} = \underset{t}{\arg\max}\dfrac{1}{n-2} \sum_{k \ne i,j} \frac{1}{\sqrt{2\pi\sigma^2}} \exp\left( -\frac{(\theta_{ij}^{(k)} - t)^2}{2\sigma^2} \right)
\end{equation}
where $\sigma := \pi/T$ controls the angular resolution. The height of the histogram peak also serves as a confidence score for the estimated dihedral angle. Compared to directly using common line angles, this voting-based aggregation improves reliability by mitigating the impact of inconsistent estimates due to noise.

\subsection{Joint Pose Estimation via Robust \texorpdfstring{$\ell_1$}{l1}-Norm Optimization}
For the rotation matrix $\boldsymbol{R}_i$, we denote its axis of rotation by $\boldsymbol{d}_i$ and the chosen X-axis of its local coordinate system (in the plane perpendicular to $\boldsymbol{d}_i$) by $\boldsymbol{q}_i$. Let $\boldsymbol{D} \in \mathbb{R}^{n \times 3}$ denote the matrix of rotation axes, and $\boldsymbol{Q} \in \mathbb{R}^{n \times 3}$ denote the matrix of in-plane X-axis vectors, one row per projection. The goal is to estimate these directions from noisy pairwise angular measurements derived from common lines. Let $\boldsymbol{\Theta} \in \mathbb{R}^{n \times n}$ denote the matrix of estimated dihedral angles between their rotation axes $\boldsymbol{d}_i$, and let $\boldsymbol{\Phi} \in \mathbb{R}^{n \times n}$ denote the matrix of angles between the in-plane local X-axis vectors. The weight matrix $\boldsymbol{W} \in \mathbb{R}^{n \times n}$ encodes the confidence in each pairwise measurement obtained as $W_{ij} :=  \underset{t}{\max}\dfrac{1}{n-2} \sum_{k \ne i,j} \frac{1}{\sqrt{2\pi\sigma^2}} \exp\left( -\frac{(\theta_{ij}^{(k)} - t)^2}{2\sigma^2} \right)$.  

Given pairwise common lines and dihedral angles, our aim is to estimate $\{\boldsymbol{R}_i\}_{i=1}^n$ through $\{\boldsymbol{d}_i\}_{i=1}^n$ and $\{\boldsymbol{q}_i\}_{i=1}^n$. In principle, this is akin to the technique of determining point coordinates from pairwise distances between the points \cite{singer2008remark}, which is the essence of the technique of multi-dimensional scaling (MDS) \cite{cox2008multidimensional}. However, for the specific problem at hand, the vectors $\{\boldsymbol{d}_i\}_{i=1}^n$ and $\{\boldsymbol{q}_i\}_{i=1}^n$ are constrained to lie on the unit sphere, so this a spherical embedding problem \cite{wilson2014}. Here, we jointly estimate $\boldsymbol{D}$ and $\boldsymbol{Q}$ by solving the following \emph{robust} optimization problem:
\begin{equation}
\begin{aligned}
\min_{\boldsymbol{D}, \boldsymbol{Q}} J(\boldsymbol{D}, \boldsymbol{Q}) := \quad & \|\boldsymbol{W} \odot (\boldsymbol{D}\boldsymbol{D}^\top - \cos(\boldsymbol{\Theta}))\|_1 \\
& \quad + \|\boldsymbol{W} \odot (\boldsymbol{Q}\boldsymbol{Q}^\top - \cos(\boldsymbol{\Phi}))\|_1 \\
\text{s.t.} \quad & \forall i, \|\boldsymbol{d}_i\|_2 = \|\boldsymbol{q}_i\|_2 = 1, \quad \boldsymbol{d}_i^\top \boldsymbol{q}_i = 0,
\end{aligned}
\label{eq:joint_opt}
\end{equation}
where $\odot$ denotes the Hadamard (element-wise) product. The cosine of in-plane angles is computed element-wise as:
\begin{equation}
\cos(\Phi_{ij}) = \cos C_{ij} \cos C_{ji} + \sin C_{ij} \sin C_{ji} \cdot (\boldsymbol{d}_i^\top \boldsymbol{d}_j),
\label{eq:phi_ij}
\end{equation}
as it is the dot product between the local X-axis vectors of $h_i$ and $h_j$ respectively given as $(\cos C_{ij}, \sin C_{ij},0)$ and $(\cos C_{ji}, \sin C_{ji} (\boldsymbol{d}_i^\top \boldsymbol{d}_j),\sin C_{ji} \sin \cos^{-1} (\boldsymbol{d}_i^\top \boldsymbol{d}_j))$.

Here, $C_{ij}$ and $C_{ji}$ (elements of the $n \times n$ matrix $\boldsymbol{C}$) are angles between the detected common lines and the chosen local X-axes of projections $i$ and $j$ respectively (computed from the particles directly). This formulation imposes exact orthogonality constraints and leverages the robustness of $\ell_1$ norms to handle outliers and inconsistencies in angular measurements, which are inevitable due to the low SNR.

\subsection{Initialization}
We adopt a two-step initialization strategy based on projected subgradient descent (PGD), enforcing unit norm constraints at each step:
\begin{enumerate}
    \item We initialize the normal vector matrix $\boldsymbol{D}^{(0)}$ by minimizing the first term in Eq.~\eqref{eq:joint_opt} using PGD, with projection onto the unit sphere after each update to enforce $\|\boldsymbol{d}_i\|_2 = 1$.
    \item Using the resulting $\boldsymbol{D}^{(0)}$, we compute $\cos(\Phi_{ij})$ from the common-line geometry via \eqref{eq:phi_ij} and initialize the in-plane vector matrix $\boldsymbol{Q}^{(0)}$ by minimizing the second term in Eq.~\eqref{eq:joint_opt}, again applying PGD with unit norm projection $\|\boldsymbol{q}_i\|_2 = 1$.
\end{enumerate}

To incentivize the orthogonality constraint $\boldsymbol{d}_i^\top \boldsymbol{q}_i = 0$, we apply a post-processing alignment step where we solve the following optimization problem:
\begin{equation}
\min_{\boldsymbol{R}^\top \boldsymbol{R} = \boldsymbol{I}} \left\| \operatorname{diag}(\boldsymbol{D}^{(0)} \boldsymbol{R} (\boldsymbol{Q}^{(0)})^\top) \right\|_2^2,
\end{equation}
to find an orthonormal matrix $\boldsymbol{R}$ and define the adjusted in-plane vectors as $\tilde{\boldsymbol{Q}}^{(0)} = \boldsymbol{Q}^{(0)} \boldsymbol{R}^\top$. The pair $(\boldsymbol{D}^{(0)}, \tilde{\boldsymbol{Q}}^{(0)})$ is then used as the initialization for the joint optimization problem where the orthogonality constraint is strictly enforced.

\subsection{Optimization via Coordinate Descent}
We solve the problem in \eqref{eq:joint_opt} using projected coordinate descent: alternating updates to $\boldsymbol{D}$ and $\boldsymbol{Q}$ with sub-gradient steps and projection onto the appropriate constraint sets.

\noindent\textbf{$\boldsymbol{D}$-update:} The gradient of the loss $J(.)$ with respect to $\boldsymbol{D}$ includes the sum of both terms from \eqref{eq:joint_opt}: one with $\boldsymbol{D}\boldsymbol{D}^\top$ and the other with $\cos (\boldsymbol{\Phi}))$ (since $\cos(\boldsymbol{\Phi})$ depends on $\boldsymbol{D}\boldsymbol{D}^\top$ as seen in \eqref{eq:phi_ij}). We define the following quantities which will help us arrive at the expression for $\nabla J_{\boldsymbol{D}}$:
\begin{align}
\boldsymbol{A} &:= \cos \boldsymbol{C} \cdot (\cos \boldsymbol{C})^\top \\
\boldsymbol{B} &:= \sin \boldsymbol{C} \cdot (\sin \boldsymbol{C})^\top \\
\boldsymbol{Z} &:= \boldsymbol{A} + \boldsymbol{B} \odot (\boldsymbol{D}\boldsymbol{D}^\top)\\
\boldsymbol{R}_{\boldsymbol{D}} &:= \boldsymbol{W} \odot (\boldsymbol{D}\boldsymbol{D}^\top - \cos(\boldsymbol{\Theta})) \\
\boldsymbol{R}_{\boldsymbol{Q}} &:= \boldsymbol{W} \odot (\boldsymbol{Q}\boldsymbol{Q}^\top - \boldsymbol{Z}).
\end{align}
This yields:
\begin{equation}
\nabla J_{\boldsymbol{D}} = 2(\boldsymbol{W} \odot \operatorname{sign}(\boldsymbol{R}_{\boldsymbol{D}}
))\boldsymbol{D} - 2(\boldsymbol{W} \odot \boldsymbol{B} \odot \operatorname{sign}(\boldsymbol{R}_{\boldsymbol{Q}}))\boldsymbol{D}.
\end{equation}
After each gradient descent update, we project each $\boldsymbol{d}_i$ to satisfy:
\begin{align}
\boldsymbol{d}_i &\leftarrow \boldsymbol{d}_i - (\boldsymbol{d}_i^\top \boldsymbol{q}_i) \boldsymbol{q}_i \quad \text{(orthogonalize)}, \\
\boldsymbol{d}_i &\leftarrow \frac{\boldsymbol{d}_i}{\|\boldsymbol{d}_i\|_2} \quad \text{(normalize)}.
\end{align}
\noindent\textbf{$\boldsymbol{Q}$-update:} Once $\boldsymbol{D}$ is updated, we recompute $\boldsymbol{Z}$ and apply the gradient descent to update $\boldsymbol{Q}$:
\begin{align}
\boldsymbol{R}_{\boldsymbol{Q}} &:= \boldsymbol{W} \odot (\boldsymbol{Q}\boldsymbol{Q}^\top - \boldsymbol{Z}) \\
\nabla J_{\boldsymbol{Q}} &= 2(\boldsymbol{W} \odot \operatorname{sign}(\boldsymbol{R}_{\boldsymbol{Q}}))\boldsymbol{Q}
\end{align}
Each $\boldsymbol{q}_i$ is similarly projected to satisfy $\boldsymbol{d}_i^\top \boldsymbol{q}_i = 0$ and $\|\boldsymbol{q}_i\|_2 = 1$.

The complete procedure is summarized in Alg.~\ref{alg:PGD}.
\begin{algorithm}
\caption{Joint Pose Estimation via Coordinate Descent}
\begin{algorithmic}[1]
\Require Initial $\boldsymbol{D}, \boldsymbol{Q}$, weight matrix $W$, learning rates $\alpha, \beta$
\For{$k = 1$ to $K_{\max}$}
    \State Compute $\boldsymbol{R}_{\boldsymbol{D}}$, $\boldsymbol{R}_{\boldsymbol{Q}}$, and gradients $\boldsymbol{\nabla}_{\boldsymbol{D}}$, $\boldsymbol{\nabla}_{\boldsymbol{Q}}$
    \State $\boldsymbol{D} \leftarrow \boldsymbol{D} - \alpha \nabla_{\boldsymbol{D}}$; enforce orthogonality and unit norm on each $\boldsymbol{d}_i$
    \State $\boldsymbol{Q} \leftarrow \boldsymbol{Q} - \beta \boldsymbol{\nabla}_{\boldsymbol{Q}}$; enforce orthogonality and unit norm on each $\boldsymbol{q}_i$
    \If{converged}
        \State \textbf{break}
    \EndIf
\EndFor
\State \textbf{return} $\boldsymbol{D}, \boldsymbol{Q}$
\end{algorithmic}
\label{alg:PGD}
\end{algorithm}

After the projection angles are estimated, the FIRM method \cite{wang2013} can be used to produce the final 3D reconstruction result.


\begin{table*}[ht]
\centering
\caption{Pose estimation errors across three synthetic maps at SNR = 0.1 using 1000 projections. Each cell reports the metric values for \textbf{Ours / SE / LUD / Sync}, where SE refers to the standard spherical embedding~\cite{lu2022}. Best result per row is in \textbf{bold}. MAE: Mean Absolute Error, MSE: Mean Squared Error.}
\label{tab:synthetic_results}
\resizebox{\textwidth}{!}{
\begin{tabular}{|l|c|c|c|}
\hline
\textbf{Metric (type)} & \textbf{EMD-3508} & \textbf{EMD-42525} & \textbf{EMD-22689} \\
\hline
$\theta_{ij}$ error (MAE) 
& \textbf{0.0267} / 0.0747 / 0.3035 / 0.2024 
& \textbf{0.00518} / 0.01008 / 0.05196 / 0.05278 
& \textbf{0.2584} / 0.3667 / 0.3874 / 0.2624 \\
\hline
$\phi_{ij}$ error (MAE) 
& \textbf{0.0196} / 0.0795 / 0.1996 / 0.1643 
& \textbf{0.00392} / 0.02995 / 0.03510 / 0.04613 
& 0.1946 / 0.3457 / 0.2676 / \textbf{0.1933} \\
\hline
In-plane rotation error (°) 
& \textbf{1.56} / 6.56 / 17.52 / 12.64 
& \textbf{0.20} / 2.01 / 1.58 / 1.94 
& 21.19 / 31.47 / 21.71 / \textbf{19.58} \\
\hline
Normal vector error (°) 
& \textbf{1.59} / 4.74 / 17.70 / 12.64 
& \textbf{0.32} / 0.60 / 2.76 / 3.14 
& 17.58 / 27.72 / 23.55 / \textbf{15.59} \\
\hline
Euler angle error $\alpha$ (MSE) 
& \textbf{0.00081} / 0.00571 / 0.05001 / 0.04878 
& \textbf{2.4e-5} / 7.0e-5 / 3.55e-3 / 1.86e-3 
& \textbf{0.0246} / 0.0559 / 0.07472 / 0.04186 \\
\hline
Euler angle error $\beta$ (MSE) 
& \textbf{0.00936} / 0.03943 / 0.35182 / 0.14755 
& \textbf{7.8e-5} / 2.6e-4 / 7.29e-3 / 5.23e-3 
& 0.6925 / 0.9536 / 0.4925 / \textbf{0.4046} \\
\hline
Euler angle error $\gamma$ (MSE) 
& \textbf{0.00930} / 0.04331 / 0.30561 / 0.16116 
& \textbf{6.4e-5} / 1.4e-3 / 3.82e-3 / 2.29e-3 
& 0.5701 / 0.7725 / 0.3833 / \textbf{0.3788} \\
\hline
\end{tabular}
}
\end{table*}

\begin{figure*}
\centering
\includegraphics[width=0.59\linewidth]{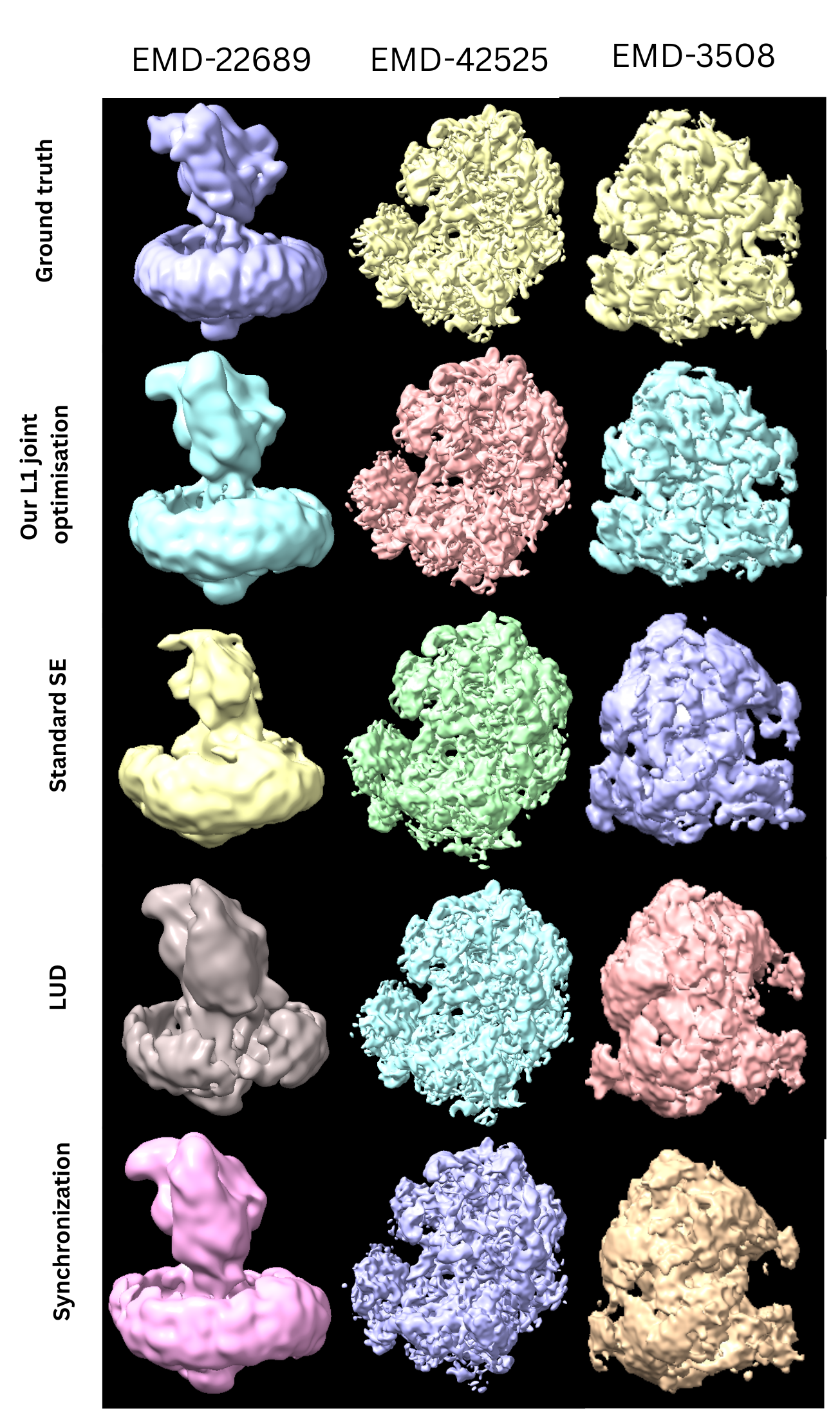}
\includegraphics[width=\linewidth]{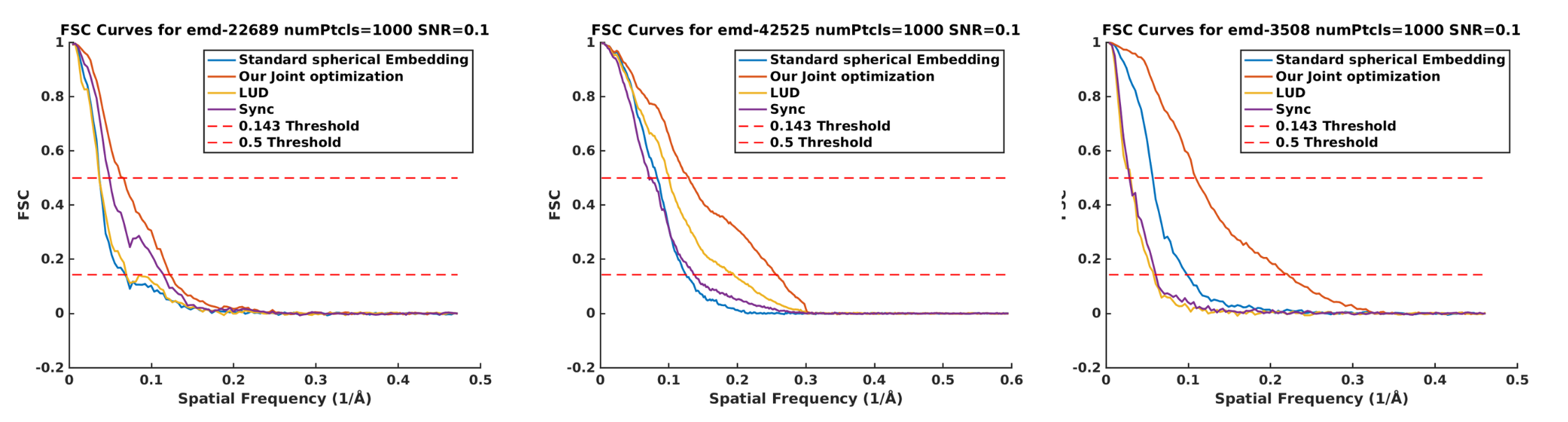}
\caption{Visual comparison and FSC curves for reconstructions using 1000 centered particles (no shift) with random orientation, each at SNR 0.1 for 3 different maps. The results confirm the efficacy of joint pose+shift estimation (cf Alg.~\ref{alg:PGD}, \ref{alg:shift_refine} and Fig.~\ref{fig:full-pipeline}) compared to competing methods: Synchronization \cite{Shkolnisky2012_sync}, LUD \cite{wang2013_lud} and SE \cite{lu2022}.}
\label{fig:joint_opt_validation}
\end{figure*}

\section{In-Plane Shift Correction via Iterative Common-Line Consistency}
In cryo-EM, accurate 3D reconstruction requires the alignment of 2D projections that may be shifted due to sample drift, beam-induced motion, or particle picking inaccuracies. Each observed projection image $\Tilde{h}_k$ is a translated version of its centered version $h_k$, with unknown in-plane shifts $(\Delta x_k, \Delta y_k)$. Neglecting these displacements can lead to significant errors in pose estimation and downstream volume reconstruction.
The work in \cite{Singer2012} introduced a method to estimate these shifts by solving a global linear system derived from pairwise common-line constraints. However, under realistic noise levels, these constraints may shift as projections are corrected, leading to inconsistent or unstable solutions. We address this by proposing an iterative refinement framework that maintains consistency between estimated shifts and detected common lines throughout the correction process.

\subsection{Shift Refinement via Common-Line Geometry}

Given a set of shifted projections, our goal is to estimate the in-plane translations such that the common-line geometry across all projection pairs becomes globally consistent. The method proceeds iteratively. In each round, the current shift estimates are used to apply phase corrections in Fourier space, which—by the Fourier shift theorem—is equivalent to translating the projections in real space. After this correction, common lines between projection pairs are re-estimated. Using the angular indices of the matched common lines, we locate the corresponding rays in the original, uncorrected Fourier slices. These rays represent the projection data before any phase correction. We then compute the relative phase offset by identifying the 1D shift that maximizes the correlation between these two rays. This ensures that the shift estimates are driven by the original data, while remaining aligned with the geometry inferred from the corrected domain. The set of estimated offsets across projection pairs forms a linear system that relates the unknown shifts to the observed phase differences. Solving this system in the least-squares sense updates the shift estimates. The process repeats until the solution converges. A formal convergence proof for this procedure is outside the scope of this work, but we have always observed numerical convergence in our experiments. The procedure for shift correction is presented in Alg.~\ref{alg:shift_refine}, which uses several notation that are defined below. 
\begin{itemize}[noitemsep, topsep=0pt]
    \item $\mathcal{F}h_k(r, \theta)$: Polar Fourier transform of $h_k$, the $k$-th particle.
    \item $\hat{\mathcal{F}}h_k(r, \theta)$: Fourier slice after phase correction of $h_k$ using current shift estimates.
    \item $(\Delta x_k, \Delta y_k)$: Unknown 2D in-plane shifts for projection $k$.
    \item $(c_i, c_j)$: Indices of common lines between two projections.
    \item $s^*$: Estimated 1D relative shift along the matched radial line.
    \item $\boldsymbol{x} \in \mathbb{R}^{2K}$: Stacked shift vector containing all $K$ projection shifts.
    \item $\boldsymbol{A} \in \mathbb{R}^{M \times 2K}$: Sparse coefficient matrix, with $M = \binom{K}{2}$ equations.
    \item $\boldsymbol{b} \in \mathbb{R}^{M}$: Observed phase offsets derived from radial-line alignment.
\end{itemize}

\begin{algorithm}
\caption{Iterative In-Plane Shift Refinement}
\label{alg:shift_refine}
\begin{algorithmic}[1]
\State \textbf{Require:} Fourier slices $\{\mathcal{F}h_k\}_{k=1}^K$; initial shifts $\{(\Delta x_k^{(0)}, \Delta y_k^{(0)})\}$ ; number of angular rays $n_\theta$
\State \textbf{Ensure:} Refined shifts $\{(\Delta x_k, \Delta y_k)\}_{k=1}^K$
\State Set convergence threshold $\epsilon$; iteration counter $t \gets 0$
\Repeat
    \State $t \gets t + 1$
    \For{$k = 1$ to $K$}
        \State Apply phase correction:
        \[
        \hat{\mathcal{F}}h_k(r, \theta) \gets \mathcal{F}h_k(r, \theta) \cdot 
        e^{-2\pi \iota r \left( \Delta x_k^{(t-1)} \sin\theta + \Delta y_k^{(t-1)} \cos\theta \right)}
        \]
        
    \EndFor

    \State Initialize sparse matrix $\boldsymbol{A}$ and vector $\boldsymbol{b}$

    \For{$k_1 = 1$ to $K-1$}
    \For{$k_2 = k_1 + 1$ to $K$}
\State Find indices $(c_{k_1}, c_{k_2})$ of common-lines in 
       $\hat{\mathcal{F}}h_{k_1}, \hat{\mathcal{F}}h_{k_2}$
\State Extract corresponding rays $r_{k_1}$, $r_{k_2}$ in $\mathcal{F}h_k$: 
\Statex \hspace{4.2em} $\mathbf{r}_{k_1} = \mathcal{F}h_{k_1}(r, \theta_{c_{k_1}})$, \quad $\mathbf{r}_{k_2} = \mathcal{F}h_{k_2}(r, \theta_{c_{k_2}})$

        \State $s^* \gets \arg\max_s \, \text{corr}(\boldsymbol{r}_{k_1} \cdot e^{-2\pi \iota s \mathbf{f}}, \boldsymbol{r}_{k_2})$
        \State $\alpha \gets \pi (c_{k_1} - 1)/n_\theta$,\quad
               $\beta \gets \pi (c_{k_2} - 1)/n_\theta$
        \State Append $[\sin\alpha,\, \cos\alpha,\, -\sin\beta,\, -\cos\beta]$ to corresponding row of $\boldsymbol{A}$ at indices $[2k_1,\, 2k_1{+}1,\, 2k_2,\, 2k_2{+}1]$ (all other entries zero)
        \State Append $s^*$ to $\boldsymbol{b}$
    \EndFor
    \EndFor

    \State Solve: $\boldsymbol{x}^{(t)} \gets \arg\min_{\boldsymbol{x}} \| \boldsymbol{A} \boldsymbol{x} - \boldsymbol{b} \|_2^2$
\Until{$\|\boldsymbol{A} \boldsymbol{x}^{(t)} - \boldsymbol{b}\|_2$ converges or $\| \boldsymbol{x}^{(t)} - \boldsymbol{x}^{(t-1)} \|$ is below $\epsilon$}
\State \Return $\boldsymbol{x}^{(t)}$
\end{algorithmic}
\end{algorithm}

\begin{figure*}
    \centering
    \includegraphics[width=\linewidth]{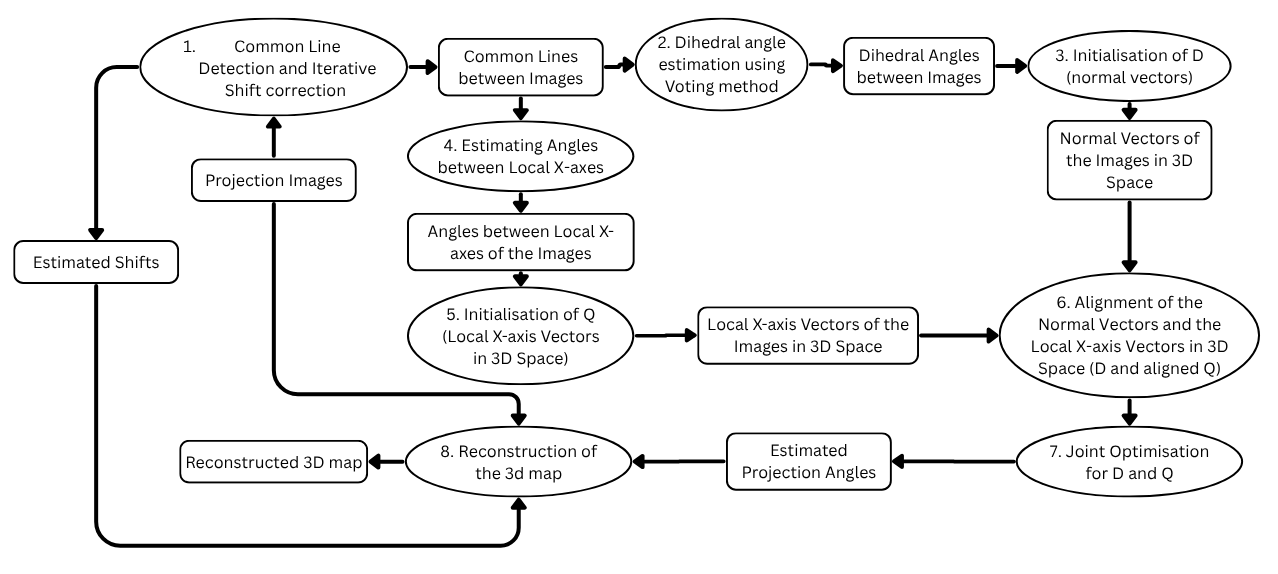}
    \caption{Full pipeline involving orientation and shift estimation}
    \label{fig:full-pipeline}
\end{figure*}
\section{Experimental Results}
\noindent\textbf{Synthetic Centered Particle Data:} We first evaluated the effectiveness of our joint $\ell_1$-norm based pose estimation on simulated cryo-EM projections corrupted with noise corresponding to a signal to noise ratio (SNR) of 0.1. In this experiment, the projections were assumed to be centered (no shifts). We simulated projections at different randomly chosen orientations of three molecular maps taken from the well known database EMDB at \url{https://www.ebi.ac.uk/emdb/}. The molecular maps were:
\begin{itemize}[noitemsep,topsep=0pt]
    \item \textbf{Map 1 (EMD-3508)} – 70S ribosome, size $260^3 \cite{emd3508_Huter2016}$
    \item \textbf{Map 2 (EMD-42525)} – 80S ribosome, size $648^3 \cite{emd42525_Musalgaonkar2025Reh1}$
    \item \textbf{Map 3 (EMD-22689)} – PTCH1 with nanobody, size $256^3 \cite{emd22689_pnas.2011560117}$
\end{itemize}
We generated 1000 projections (corrupted by additive white Gaussian noise) per map and compared our method to the spherical embedding approach (SE) ~\cite{lu2022}, least unsquared deviations (LUD)~\cite{wang2013_lud}, and synchronization-based pose estimation (Sync) ~\cite{Shkolnisky2012_sync}. Table~\ref{tab:synthetic_results} reports errors for dihedral and in-plane angle estimation, as well as errors for in-plane rotation and Euler angles. Fig.~\ref{fig:joint_opt_validation} presents visual comparisons of the reconstruction and Fourier Shell Correlation (FSC) curves for three synthetic maps. The FSC curve is a plot of the correlation values between Fourier coefficients of the two maps being compared, as a function of the 2D frequency magnitude. The `taller' the plot (higher the correlation), the better is the reconstruction fidelity. We observe that reconstructions generated using our joint optimization framework align more closely with the ground truth, both in terms of structural detail and frequency preservation, as indicated by higher FSC values across resolutions. In particular, for Map 3 (EMD-22689), it is clearly visible in Fig.~\ref{fig:joint_opt_validation} that our method yields the most accurate 3D reconstruction among all baselines. These results confirm that robust joint pose estimation not only improves angular consistency but also leads to significantly better reconstructions, even under high noise conditions. The `ground truths' reported in Fig.~\ref{fig:joint_opt_validation} correspond to RELION/Cryosparc reconstructions with 30,000 to 300,000 particles (depending on the dataset), and our technique achieves high fidelity reconstructions with just 1000 particles. 

\begin{figure*}
\centering
\includegraphics[width=0.65\linewidth]{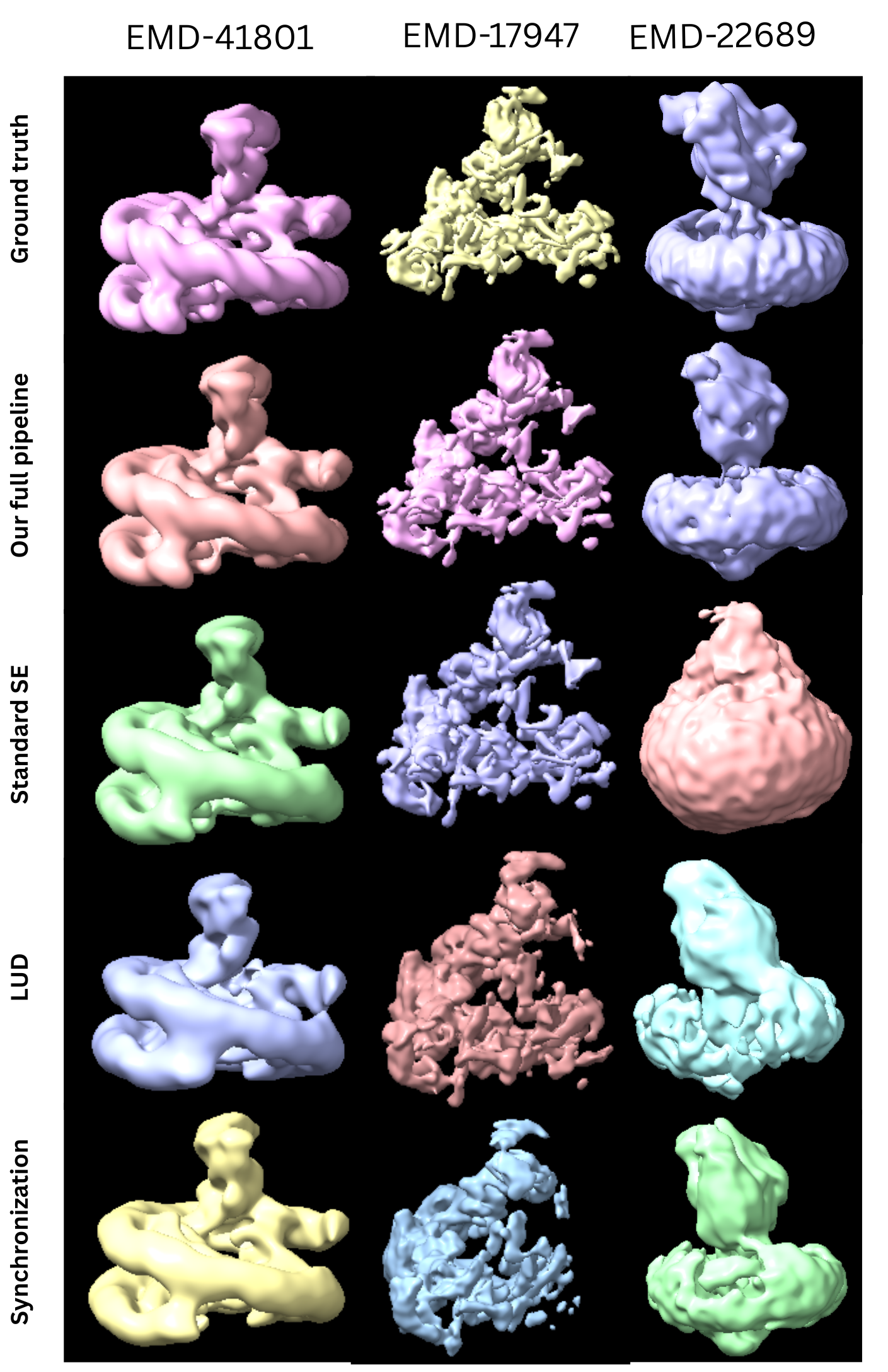}
\includegraphics[width=\linewidth]{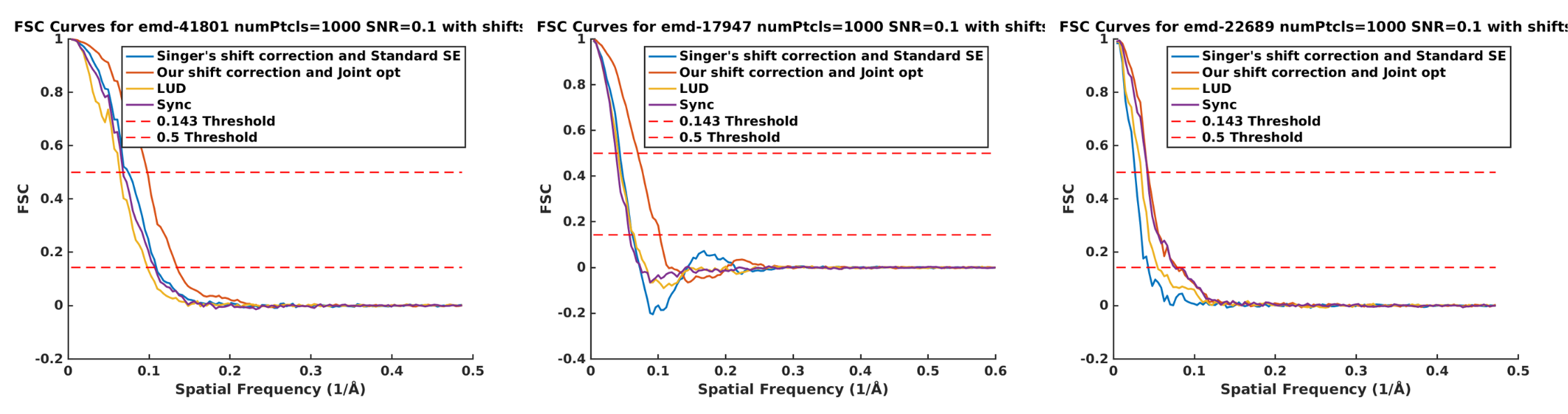}
\caption{Visual comparison and FSC curves for reconstructions using 1000 particles with random orientation and shift, each at SNR 0.1 for 3 different maps. The results confirm the efficacy of joint pose+shift estimation (cf Alg.~\ref{alg:PGD}, \ref{alg:shift_refine} and Fig.~\ref{fig:full-pipeline}) compared to competing methods: Synchronization \cite{Shkolnisky2012_sync}, LUD \cite{wang2013_lud} and SE \cite{lu2022}.}
\label{fig:full_pipeline_validation_synthetic}
\end{figure*}

\noindent\textbf{Synthetic Particle Data With Shifts:} We then evaluated our complete pipeline—including both pose estimation and shift correction, shown in Fig.~\ref{fig:full-pipeline}—on three derived molecular maps:
\begin{itemize}[noitemsep,topsep=0pt]
    \item \textbf{Map 1 (EMD-41801)} – Cryo-EM structure of the human nucleosome core particle ubiquitylated at histone H2A lysine 15 in complex with RNF168-UbcH5c (class 2), size $256^3 \cite{emd41801_Qi2024}$
    \item \textbf{Map 2 (EMD-17947)} – Trimeric prM/E spike of Tick-borne encephalitis virus
immature particle, size $300^3 \cite{emd_17947_Maria2024}$
    \item \textbf{Map 3 (EMD-22689)} – PTCH1 with nanobody, size $256^3 \cite{emd22689_pnas.2011560117}$
\end{itemize}
For each map, we simulated 1000 shifted projections corrupted with Gaussian noise, again corresponding to an SNR of 0.1. We compared our full pipeline (cf Fig.~\ref{fig:full-pipeline}) against a baseline shift correction method combined with three existing angle estimation techniques: spherical embedding (SE)~\cite{lu2022}, the least unsquared deviations (LUD) formulation~\cite{wang2013_lud}, and the synchronization method~\cite{Shkolnisky2012_sync}. As described in Fig.\ref{fig:full_pipeline_validation_synthetic}, our full pipeline—combining robust angle estimation with an improved shift correction procedure—produces reconstructions that exhibit greater structural detail and fidelity to the ground truth. Notably, for Map 1 (EMD-41801), our method preserves fine features more accurately than competing approaches.  Again, the `ground truths' reported in Fig.~\ref{fig:full_pipeline_validation_synthetic} correspond to RELION/Cryosparc reconstructions with 30,000 to 300,000 particles, and our technique achieves high fidelity reconstructions with just 1000 particles.
 
\begin{figure*}[ht]
    \centering
    \includegraphics[width=0.3\linewidth]{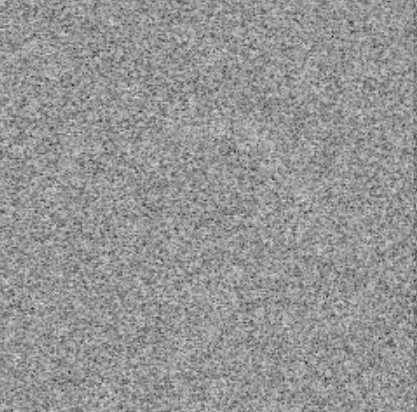}
    \includegraphics[width=0.3\linewidth]{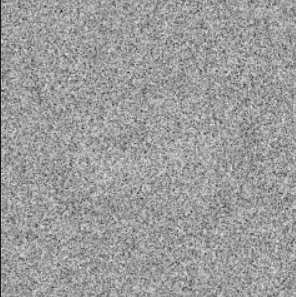}
    \caption{Example micrograph of \textit{Escherichia coli} 70S Ribosome: the projections in the micrograph exhibit extremely low signal-to-noise ratio (SNR), making it challenging for pose estimation and reconstruction algorithms—especially those designed to operate with a limited number of particles—to recover meaningful 3D structure.}
    \label{fig:sample_raw_particles}
\end{figure*}

\begin{figure*}[ht]
    \centering
    \includegraphics[width=0.4\linewidth]{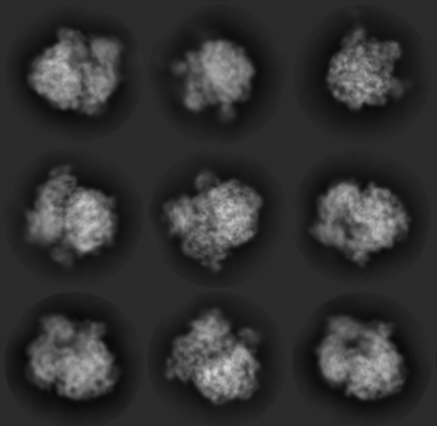}
    \caption{Sample class averages generated by RELION starting from particles extracted from the same micrograph of the 70S Ribosome. Each class average represents the mean of thousands of similar particle images, substantially improving the SNR. These noise-reduced projections enable accurate 3D reconstructions even with a small number of input views. However, generating class averages requires acquiring and processing millions of raw particles.}
    \label{fig:sample_class_averages}
\end{figure*}

\noindent\textbf{Experiments with Class Average Data:} For real datasets acquired in the form of micrographs, ground-truth 3D maps are not available. To serve as a reference for evaluation, we used high-quality reconstructions obtained from the RELION pipeline, which processes millions of particle images to produce state-of-the-art maps. To mitigate the extreme noise levels inherent in raw single-particle images (as seen in Fig.~\ref{fig:sample_raw_particles}), we used class-averaged 2D projections as input to all methods. This preprocessing step substantially improves the signal-to-noise ratio (SNR) (as shown in Fig.~\ref{fig:sample_class_averages}), although it can potentially lead to loss of spatial resolution due to the creation of averages.

\begin{figure*}
\centering
\includegraphics[width=0.5\linewidth]{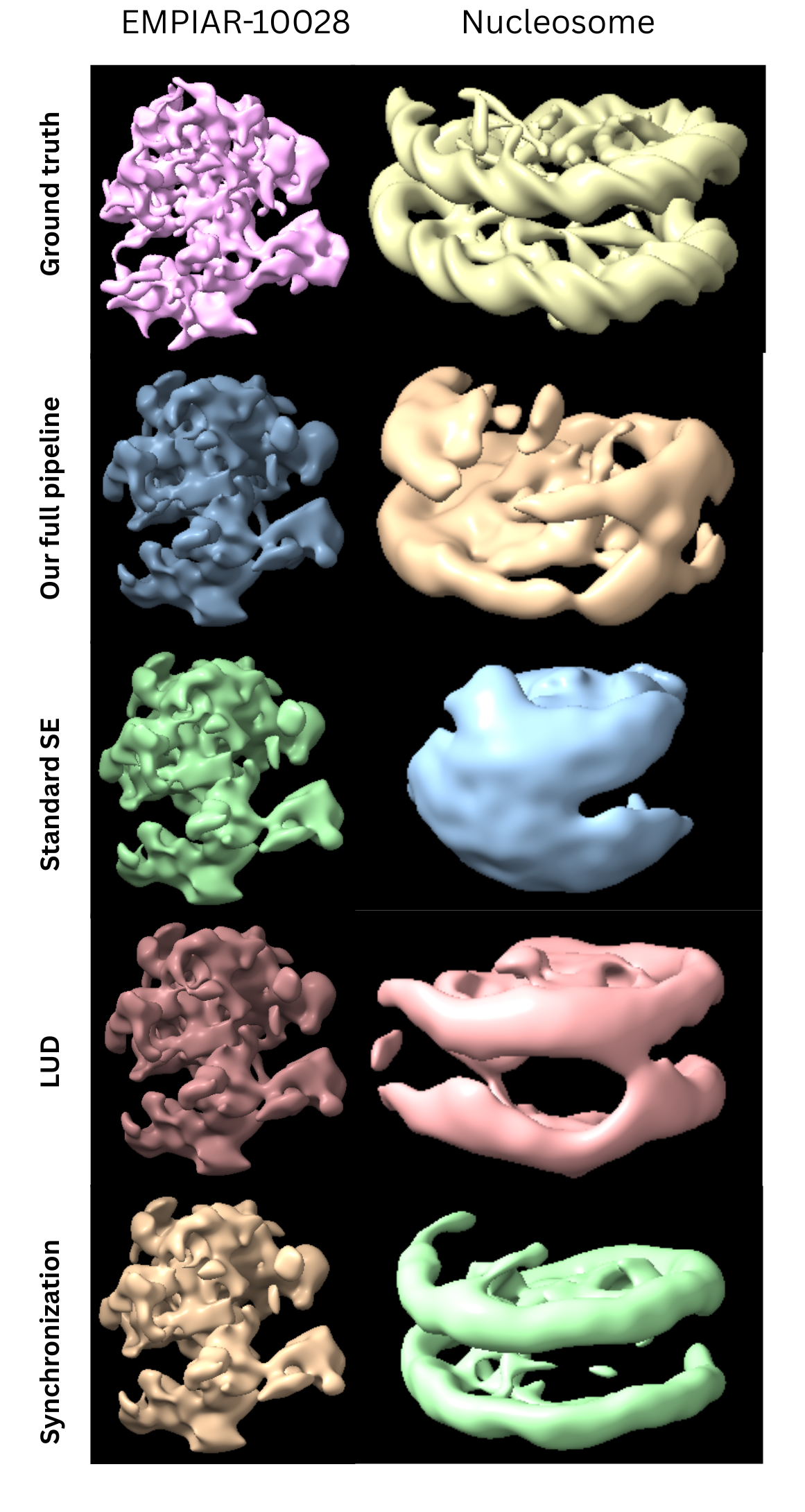}
\includegraphics[width=0.6\linewidth]{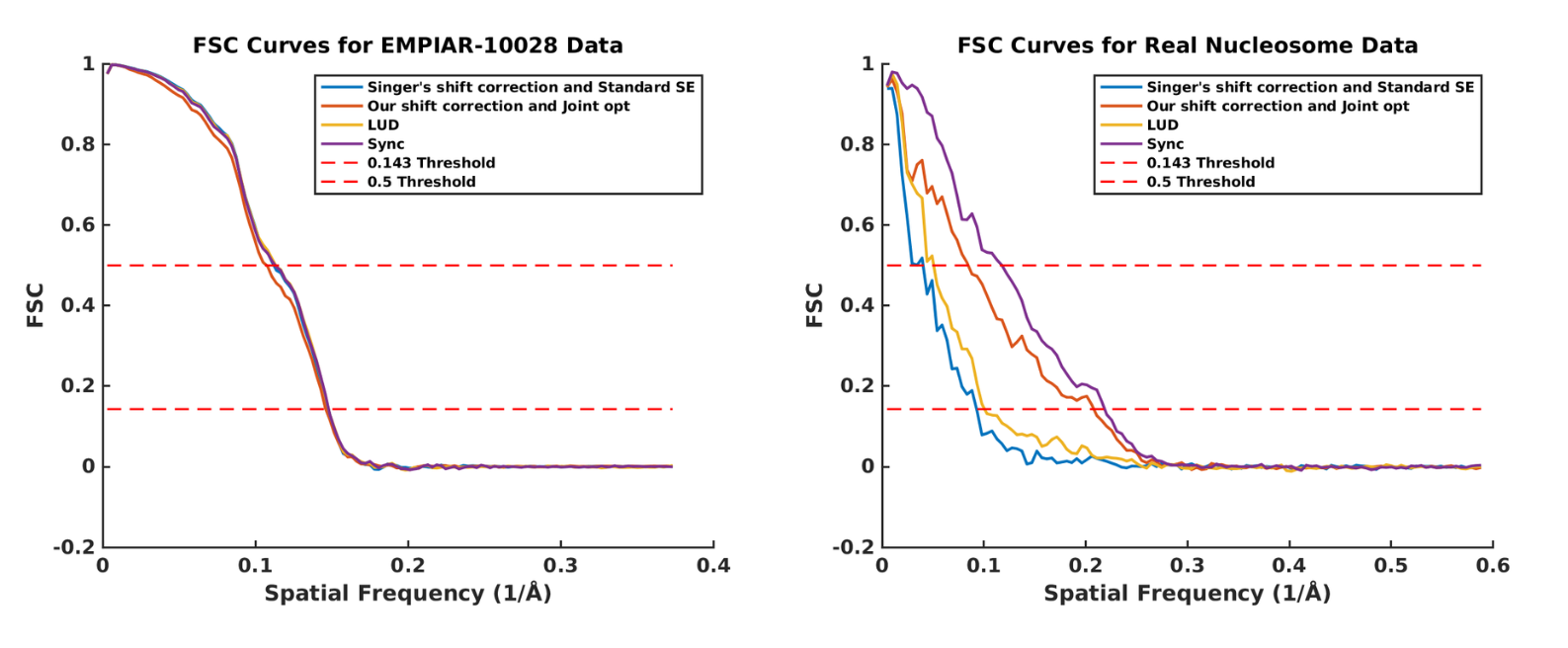}
\caption{Visual comparison and FSC curves for two real datasets, algorithms were run on class averages created from millions of particles.}
\label{fig:full_pipeline_validation_real}
\end{figure*}

We tested our pipeline on two distinct datasets:
\begin{itemize}[noitemsep,topsep=0pt]
    \item \textbf{Plasmodium falciparum 80S ribosome} bound to the anti-protozoan drug emetine, consisting of approximately 600 class averages~\cite{wong_empiar_10028}
    \item \textbf{Nucleosome particle}, with 851 class averages
\end{itemize}
As shown in Fig.\ref{fig:full_pipeline_validation_real}, most methods --including our proposed approach-- produce high-quality reconstructions across datasets. An exception is observed in the nucleosome dataset, where the method of \cite{lu2022} fails to recover a meaningful structure. Overall, the improved SNR due to class averaging allows even baseline methods to perform competitively, highlighting the importance of using more challenging low-SNR inputs to rigorously evaluate pose estimation robustness.

\noindent\textbf{Discussion:} Numerical results apart, our technique has some distinct principled advantages over previous techniques. The techniques in \cite{lu2022} and \cite{wang2024orientation} do not enforce a strict orthogonality on the rotation matrices during the optimization and compute the matrices via an external projection step after the optimization. The method in \cite{Shkolnisky2012_sync} is based on the computation of the eigenvectors of the synchronization matrix, which is susceptible to errors in dihedral angles induced by noise in common lines (see Equations (4.12) to (4.17) of \cite{Shkolnisky2012_sync}). The technique in \cite{wang2013_lud} is more noise robust due to the use of unsquared deviations. However, it does not inherently impose orthogonality of the off-diagonal blocks of the Gram matrix (see equations (4.3) to (4.9) of \cite{wang2013_lud}). 

\section{Conclusion and Future Work}
In this paper, we introduced a robust joint optimization framework for cryo-EM pose estimation that improves upon existing methods in both angular alignment and in-plane shift correction. Our first contribution is a principled $\ell_1$-norm based spherical embedding algorithm that accurately estimates projection parameters under high noise conditions, outperforming state-of-the-art approaches such as LUD~\cite{wang2013_lud}, synchronization-based method~\cite{Shkolnisky2012_sync}, and the spherical embedding method \cite{lu2022}. The second contribution enhances the shift correction procedure proposed by \cite{Singer2012} through an iterative refinement strategy that enforces consistency between the common lines and shifts. Our full pipeline, illustrated in Fig.~\ref{fig:full-pipeline}, consistently yields more accurate 3D reconstructions across synthetic datasets. However, a key limitation lies in the reliability of common lines created in the iterative shift refinement. In particular, under extreme noise, the common-line matches may degrade in quality compared to those generated by \cite{Singer2012}. Future work will focus on improving the reliability of common-line detection within our iterative framework and further enhancing robustness to noise. The pipeline proposed in this paper offers the significant advantage of enabling real-time 3D reconstruction of biomolecules during the costly cryo-EM data collection process. This on-the-fly 3D reconstruction empowers researchers to make immediate and informed decisions regarding sample quality, identify potential ligand binding sites on the biomolecules, and achieve high-resolution structural insights, thereby greatly enhancing the efficiency and effectiveness of cryo-EM data collection and data  processing.

{
    \small
    \bibliographystyle{ieeenat_fullname}
    \bibliography{main}
}

\end{document}